\documentclass[runningheads]{llncs}
\usepackage{graphicx}
\usepackage{xr-hyper} 
\usepackage{hyperref}

\usepackage{amssymb}
\usepackage{amsmath}
\usepackage{enumerate} 
\usepackage{pifont} 
\usepackage{algpseudocode}
\usepackage{algorithm}
\algtext*{EndWhile}
\algtext*{EndIf}
\algtext*{EndFor}
\algtext*{EndFunction}
\algtext*{EndProcedure}
\usepackage[utf8]{inputenc}
\usepackage[english]{babel}

\usepackage{graphicx}
\usepackage{xcolor}
\graphicspath{{figures/}}
\usepackage{subcaption}
\usepackage{sidecap}
\usepackage{wrapfig}
\usepackage{multirow}
\usepackage{booktabs}
\usepackage{placeins} 
\usepackage{dsfont} 

\usepackage{cleveref} 

\newcommand{\etal}{\textit{et al}.}
\newcommand{\ie}{\textit{i}.\textit{e}.}

\newcommand{\norm}[1]{\left\lVert#1\right\rVert}

\newcommand{\highlight}[1]{{\color{blue}\textbf{#1}}}

\begin{document}
	\title{Adaptive Learning Rate and Momentum for Training Deep Neural Networks}
	\titlerunning{Ada. LR and Momentum for Training DNNs}
	%
	\author{
		Zhiyong Hao\inst{1} \and
		Yixuan Jiang\inst{1} \and
		Huihua Yu\inst{1} \and
		Hsiao-Dong Chiang\inst{1}
	}
	%
	\authorrunning{Z. Hao et al.}
	%
	\institute{
		Cornell University, Ithaca NY 14850, USA\\
		\email{\{zh272,yj373,hy437,hc63\}@cornell.edu}
	}
	\maketitle              
	\begin{abstract}
		Recent progress on deep learning relies heavily on the quality and efficiency of training algorithms. In this paper, we develop a fast training method motivated by the nonlinear Conjugate Gradient (CG) framework. 
		We propose the Conjugate Gradient with Quadratic line-search (CGQ) method. On the one hand, a quadratic line-search determines the step size according to current loss landscape. On the other hand, the momentum factor is dynamically updated in computing the conjugate gradient parameter (like Polak-Ribiere).
		Theoretical results to ensure the convergence of our method in strong convex settings is developed. And experiments in image classification datasets show that our method yields faster convergence than other local solvers and has better generalization capability (test set accuracy). One major advantage of the paper method is that tedious hand tuning of hyperparameters like the learning rate and momentum is avoided.
		
	\end{abstract}
	%
	%
	%
	\section{Introduction}
	We consider the minimization problem commonly used in the machine learning setting with the following finite-sum structure:
	\begin{equation}
		\underset{\theta}{\min}~f(\theta)=\frac{1}{n}\sum_{i=1}^n f_i(\theta)
		\label{eq:optim1}
	\end{equation}
	where $f:D\subset\mathds{R}^n\to\mathds{R}$ is continuously differentiable and $D$ is the search space. A point $\theta^*\in D$ is called a local minimum if $f(\theta^*)\leq f(\theta)$ for all $\theta\in D$ with $\norm{\theta-\theta^*}<\sigma$ for $\sigma>0$.  
	A general training procedure is in \Cref{alg:opt_iter}, where key steps are determining the update direction $p_t$ and step size $\alpha_t$ in each iteration.
	
	\begin{algorithm}
		\caption{Iterative Approach for Optimizing (\ref{eq:optim1}) }\label{alg:opt_iter}
		\begin{algorithmic}[1]
			\Procedure {\texttt{Optimize}}{$f(\theta), maxiter$}
			\State {Initialize $\theta_0$}
			\For{$t \gets 1$ to $maxiter$}
			\State Choose a search direction $p_t$;
			\State Determine a update step size $\alpha_t$;
			\State $\theta_t \gets \theta_{t-1}+\alpha_t p_t$;
			\EndFor
			\State \textbf{return} $\theta_t$
			\EndProcedure
		\end{algorithmic}
	\end{algorithm}
	
	Several methods propose various choices of search directions. The Gradient Descent (GD) method directly takes the negative gradient $-\nabla_\theta~f(\theta)$. Newton's method selects $p_t$ as $-(\nabla^2_\theta~f(\theta))^{-1}\nabla_\theta f(\theta)$ (where $\nabla^2_\theta f(\theta)$ is the Hessian matrix). On the other hand, momentum~\cite{polyak,nesterov} is used to stabilize $p_t$ in stochastic optimization. Adam~\cite{adam} formalizes $p_t$ as a moving average of first moments divided by that of second moments. 
	Other methods focus on the choice of $\alpha_t$. Various classic line-search methods were proposed, like the Newton-Raphson, Armijo, and Wolfe line search. Recently, \cite{painless} applied the Armijo rule in SGD. In this paper, we propose the Conjugate Gradient with Quadratic line-search method that jointly optimizes both $p_t$ and $\alpha_t$ for fast convergence.
	
	The performance of SGD is well-known to be sensitive to step size $\alpha_t$ (or the learning rate called in machine learning). Algorithms, such as AdaGrad~\cite{adagrad} and Adam~\cite{adam}, ease this limitation by tuning the learning rate dynamically using past gradient information. Line-search strategy is another option to adaptively tune step size $\alpha_t$. Exact line-search requires solving a simpler sub-optimization problem, such as applying quadratic interpolation~\cite{Powell64}. An inexact line-search method can also be applied, such as the Armijo rule~\cite{armijo1966}, to guarantee a sufficient decrease at each step. A stochastic variant of Armijo rule was proposed in \cite{painless} to set the step size for SGD.
	
	Conjugate Gradient method (CG) is also popular for solving nonlinear optimization problems. In CG, $p_t$ is chosen as $(-\nabla f(\theta)+\beta_t p_{t-1})$, where $\beta_t$ is the conjugate parameter computed according to various formulas.  $\alpha_t$ is determined by a line search ($\arg\min_\alpha f(\theta_t+\alpha p_t)$). In fact, CG can be understood as a Gradient Descent with an adaptive step size and dynamically updated momentum. For the classic CG method, step size is determined by the Newton-Raphson method or the Secant method, both of which need to exactly compute or approximate the Hessian matrix. This makes it less appealing to apply classic CG in deep learning.
	
	In this paper, we propose the Conjugate Gradient with the Quadratic line-search method (CGQ), which uses quadratic interpolation line-search to choose the step size and adaptively tunes momentum term according to the Conjugate Gradient formula. CGQ requires no Hessian information and can be viewed as a variant of SGD with adaptive learning rate and momentum. For quadratic line-search, we propose two variants, \ie, the 2-point method and the least squares method, which are introduced in detail in \Cref{sec:2pt,sec:ls}. And we illustrate how we dynamically determine the momentum term during training in \Cref{sec:polak}.
	Our analyses indicate that SGD with 2-point quadratic line-search converges under convex and smooth settings. sCGQ is further proposed to improve efficiency and generalization on large models. Through various experiments, CGQ exhibits fast convergence, and is free from manually tuning the learning rate and momentum term. 
	
	Our major contributions are summarized as follows:  
	\begin{itemize}
		\item A quadratic line-search method is proposed to dynamically adjust the learning rate of SGD.
		\item An adaptive scheme for computing the effective momentum factor of SGD is designed.
		\item Theoretical results are developed to prove the convergence of SGD with quadratic line-search.
		\item Line search on subset of batches is proposed to improve efficiency and generalization capability.
	\end{itemize}
	
	\section{Related Work}
	In modern machine learning, SGD and its variants~\cite{adagrad,adam,adamw,reddi2018}, are the most prevalent optimization methods used to train large-scale neural networks despite their simplicity. 
	Theoretically, the performance of SGD with a constant learning rate has been analyzed that under some assumptions, such as, the Strong Growth Condition, has linear convergence if the learning rate is small enough~\cite{schmidt2013fast}. However, there are still two non-negligible limitations: (1) the theoretical convergence rate is not as satisfying as the full gradient descent method; and (2) the performance is very sensitive to hyperparameters, especially the learning rate.

	In SGD, random sampling introduces randomness and degrades the theoretical convergence of SGD. One direction of research focuses on reducing the variance of random sampling so that larger learning rates are allowed. Based on this idea, a family of SGD variants called Variance Reduction methods were proposed~\cite{sag2012,svrg2013,sdca2013,saga2014}. These methods have been proved to possess impressive theoretical convergence on both convex and non-convex optimization problems. However, the applications of these methods in machine learning are limited to optimization of simple logistic regression models because the additional memory and computational cost of these methods are unacceptable when training large-scale deep neural networks.

	On the other hand, significant efforts seek to ameliorate SGD by designing an effective method to adaptively tune the learning rate as training.  \cite{pesme2020} divides the optimization process of SGD into transient phase and stationary phase, while the learning rate will be reduced after entering the stationary phase. In \cite{smith2017cyclical}, scheduling learning rate cyclically for each iteration empirically boosts the convergence of SGD. Meta-learning~\cite{baydin2018online} techniques that optimize the learning rate according to the hyper-gradient also have an outstanding performance in experiments. However, none of these techniques gives solid theoretical proof of fast convergence. In pursuit of an optimal learning rate schedule with a theoretical convergence guarantee, Vaswani \etal hybrid SGD with classic optimization techniques like Armijo line-search~\cite{painless} and Polyak step size~\cite{Loizou2020,Berrada2019}, and the developed methods not only have sound theory foundation but also empirically show much faster convergence than many widely-used optimization methods. Our work is inspired by Quadratic Interpolation (QI) technique used in line-search~\cite{qi2017}. We developed two versions of quadratic line-search method to assist SGD to automatically determine an ideal learning rate schedule. 
	
	To accelerate convergence of SGD, heavy ball algorithm (SGD with fixed momentum)~\cite{polyak} is proposed . In practical neural network training, Nesterov accelerated momentum~\cite{nesterov} is often used. As discussed in \cite{Bhaya2004}, heavy ball method is highly related to conjugate gradient (CG) method, and many efforts have been put into improving SGD by injecting the spirit of CG method. CG method presents promising performance in training an auto-encoder comparing with SGD and LBFGS~\cite{le2011}. A CG-based ADAM~\cite{Kobayashi2020} was proposed and showed comparative performance to Adam, AdaGrad in solving text and image classification problems, but no generalization performance was provided. Jin \etal~\cite{Jin2019} proposed a stochastic CG with Wolfe line-search that was tested on low-dimensional datasets. All of these related works motivate us to build a bridge between SGD and CG method and create a dynamic acceleration scheme for SGD. Instead of tuning momentum term manually, we dynamically adjust it during training according to Polak-Ribiere formula ~\cite{polak1969} which shows the best based on our experiments. 
	
	Most recently, \cite{pal2020} independently proposed a line search method based on parabolic approximation (PAL). Our major differences are: 1). CGQ introduces feedback to check the estimation quality that improves reliability. 2). CGQ eliminates the momentum hyperparameter, as discussed in \Cref{sec:polak}; 3). For stochastic settings, CGQ does line search on a subset of batches to improve the runtime and generalization; 4). Convergence analysis of CGQ is conducted with no quadratic objective assumption. 
	
	\section{The CGQ Method}
	\subsection{Overview}
	The Conjugate Gradient with Quadratic Line-Search (CGQ) method follows the iterative training framework in \Cref{alg:opt_iter}, and eliminates hand-tuning two major hyperparameters in SGD: the learning rate and momentum factor.
	
	In each training iteration, CGQ optimizes the learning rate $\alpha_t$ by estimating the one-dimensional loss landscape as a parabola. Two variants, the 2-point quadratic interpolation and the Least Squares estimation, are illustrated in \Cref{sec:lr}.
	For direction $p_t$, methods from Conjugate Gradient are adopted to dynamically adjust the momentum factor in SGD instead of using a hand-tuned fixed momentum, which is described in \cref{sec:polak}.
	
	\subsection{Dynamically Adjusting the Learning Rate}\label{sec:lr}
	\subsubsection{Two-point Quadratic Interpolation}\label{sec:2pt}
	\begin{figure*}[t]
		\centering
		\begin{subfigure}[b]{0.48\textwidth}
			\includegraphics[width=\textwidth,height=0.7\textwidth]{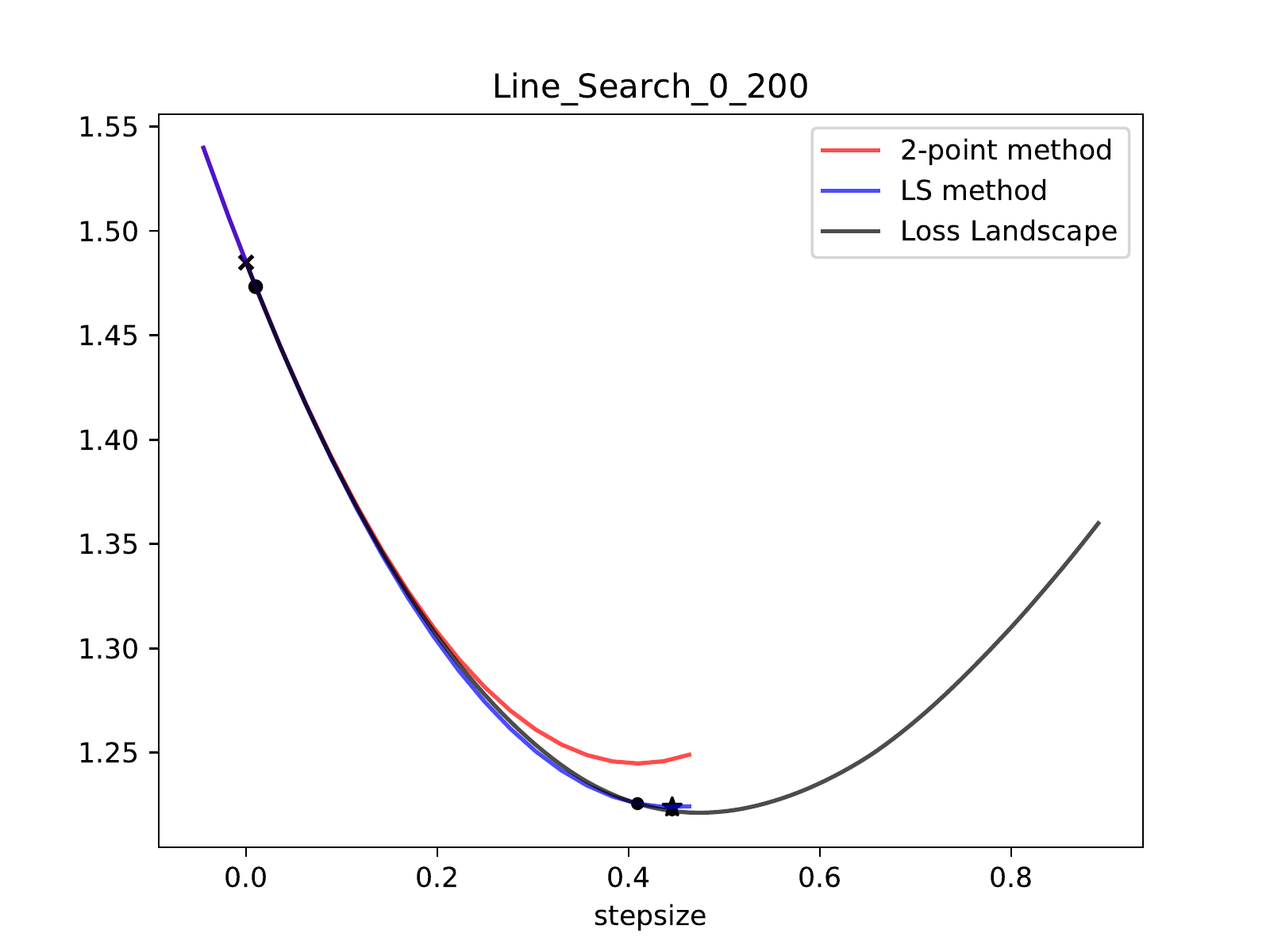}
			\caption{}
			\label{fig:smooth_est}
		\end{subfigure}
		\hfil
		\begin{subfigure}[b]{0.48\textwidth}
			\includegraphics[width=\textwidth,height=0.7\textwidth]{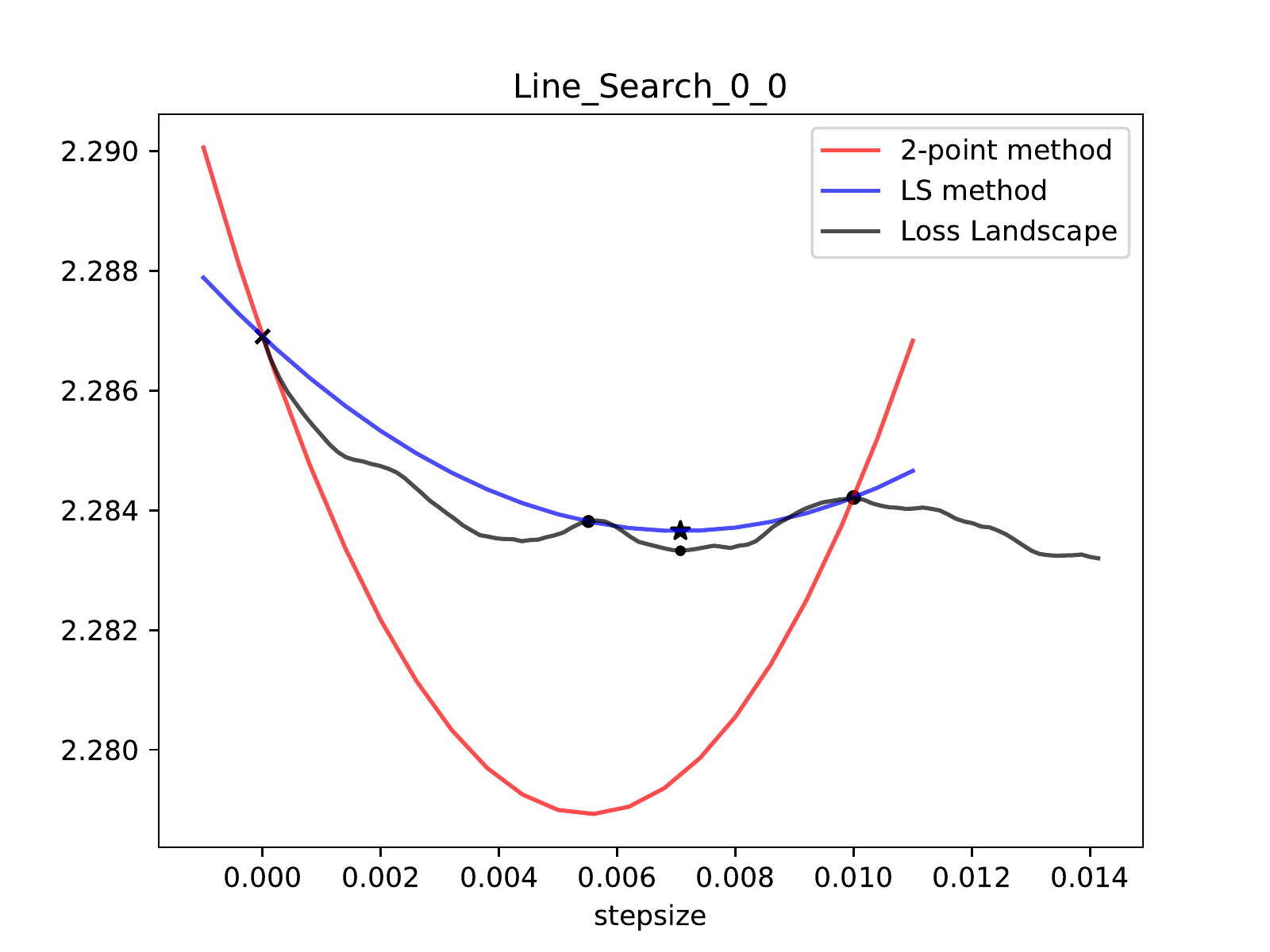}
			\caption{}
			\label{fig:rough_est}
		\end{subfigure}
		\caption{\small Comparison of the two quadratic estimation algorithms (2-point method and least squares method). (a) When the landscape is smooth, both estimate well. (b) When the landscape is rough, least squares tend to capture more details. It is also observed in experiments that roughness only occurs at the beginning of a training, after which the landscapes are almost always like (a).}
		\label{fig:landscape}
	\end{figure*}
	
	Given a twice differenciable loss function $f(\theta)$ ($\theta\in\mathds{R}^m$ is the parameter vector), its second order Taylor expansion around $\theta_0$ is expressed as:
	\begin{equation}
		\begin{split}
			f(\theta)&=f(\theta_0)+\nabla f(\theta_0)^T(\theta-\theta_0)\\
			&~~~~+\frac{1}{2}(\theta-\theta_0)^T\nabla^2f(\theta_0)(\theta-\theta_0)\\
			&~~~~+O((\theta-\theta_0)^2).
		\end{split}
	\end{equation}
	
	Along direction $p_t\in\mathds{R}^m$, the 1D loss landscape $\phi(\alpha)$ w.r.t. the step size $\alpha\in\mathds{R}$ can be described as:
	\begin{equation}
		\begin{split}
			\phi(\alpha)&=f(\theta-\alpha p_t)\\
			&=f(\theta_0)-\alpha\cdot~\nabla f(\theta_0)^Tp_t\\
			&~~~~+\frac{1}{2}\alpha^2\cdot p_t^T\nabla^2 f(\theta_0)p_t+O(\alpha^2).\\
		\end{split}
	\end{equation}
	When the residual term $O(\alpha^2)$ is small enough, the line-search function can be expressed as:
	\begin{equation}
		\phi(\alpha)\approx q(\alpha)\triangleq~A\alpha^2+B\alpha+C
	\end{equation}
	where $A=\frac{1}{2}p_t^T\nabla^2 f(\theta_0)p_t$, $B=-\nabla f(\theta_0)^Tp_t$, and $C=f(\theta_0)$.
	When $A>0$, $q(\alpha)$ reaches minimum at $\alpha^*=-B~/~(2A)$.
	
	To get $A,B,C$ at each training iteration without evaluating the Hessian matrix $\nabla^2 f(\theta_0)$, we propose to apply the 2-point quadratic interpolation method. Given two sample points $(0, q(0))$, $(\alpha_1,q(\alpha_1))$ and the slope $q'(0)$, parabola parameters can be calculated as:
	\begin{equation}
		\begin{split}
			A&=\frac{q(\alpha_1)-q(0)-q'(0)\cdot \alpha_1}{\alpha_1^2}\\
			B&=q'(0)\\
			C&=q(0)\\
		\end{split}
	\end{equation}
	The stopping criteria and details are shown in \Cref{alg:qls}.
	
	\begin{algorithm}
		\small
		\caption{Quadratic Line-Search (2-point interpolation)}\label{alg:qls}
		\begin{algorithmic}[1]
			\Procedure {\texttt{QLS2}}{$f(\cdot), \theta_t, p_t , slope$}
			\State {Initialize: $\gamma,K_{max},\alpha_{max},\sigma$}
			\State{$(x_0,y_0)\gets (0,f(\theta_t))$}
			\For{$k \gets 1$ to $K_{max}$}
			\State{$(x_1,y_1)\gets (\gamma,f(\theta_t+\gamma\cdot p_t))$}
			\State{$q_{A,B,C}(\cdot)\gets \texttt{Interpolate}(x_0,y_0,x_1,y_1,slope)$}
			\State{$\alpha^*=\min(-B~/~2A,\alpha_{max})$}
			\If{$A<=0$}
			\State{\textbf{return} $\gamma$}
			\ElsIf{$q_{ABC}(\alpha^*)>0 ~\&~ f(\theta_t+\alpha^* p_t)<f(\theta_t)$}
			\State{\textbf{return} $\alpha^*$}
			\Else{}
			\State{$\gamma\gets\alpha^*$}
			\EndIf
			\EndFor
			\State \textbf{return} $\alpha^*$
			\EndProcedure
		\end{algorithmic}
	\end{algorithm}
	
	One of the major advantages of our method is that when moving closer to a local optimum as training proceeds, $\alpha^*$ will decay automatically, due to the nature of a convex parabola. This is also verified and visualized in our experiment section. 
	Another issue is when the initial $\gamma$ is too large compared to the estimated $\alpha^*$. In such cases, $q(\alpha)$ is not a good surrogate to represent the local landscape of $\phi(\alpha)$ near the origin. This is handled by decaying the initial value of $\gamma$ when it is consistently larger than the recent average of $\alpha^*$. Through experiments, we observe that $A$ is usually estimated positive in training neural networks, while the $A<0$ case is also properly dealt with in \Cref{alg:qls}.
	
	\subsubsection{Convergence Analysis}\label{sec:convergence}
	Our method is majorly motivated by empirical results. To give some insights on the theoretical side, we provide a convergence analysis of quadratic line-search under SGD update rule. To start with, we use the following assumptions on the loss function $f(\theta)=\sum_i f_i(\theta)$, which are commonly used for convergence analysis in deep learning \cite{painless,Loizou2020}.
	(a) $f$ is \textit{lower-bounded} by a finite value $f^*$;
	(b) \textit{interpolation}: This requires that $\nabla f(\theta^*)=0$ implies $\nabla f_i(\theta^*)=0$ for all components $f_i$;
	(c) $f$ is \textit{$L$-smooth}, \ie, the gradient $\nabla f$ is $L$-Lipschitz continuous; and
	(d) $f$ is \textit{$\mu$ strong-convex}.
	Proofs are given in the appendix.
	
	The following lemma provides the bound of the step size returned by quadratic line-search at each iteration $k$.
	\begin{lemma}
		Assuming 
		(1) $f_i$'s are $L_i$-smooth, 
		(2) $\mu_i$ strong-convexity of $f_i$, 
		the step size $\alpha_k$ returned by quadratic line-search in \cref{alg:qls} that is constrained within $(0,\alpha_{max}]$ satisfies:
		\begin{equation}
			\alpha_k \in [\min\{\frac{1}{L_{ik}}\text{, }\alpha_{max}\}\text{, } \min\{\frac{1}{\mu_{ik}}\text{, }\alpha_{max}\}].
		\end{equation}
		\label{lemma:stepsize}
	\end{lemma}
	
	Next, we provide the convergence of quadratic line-search under strong-convex and smooth assumptions.
	\begin{theorem}
		Assuming 
		(1) interpolation, 
		(2) $f_i$'s are $L_i$-smooth, and
		(3) $\mu_i$ strong-convexity of $f_i$'s.
		When setting $\alpha_{max} \in (0,~\min_i\{\frac{1}{L_i} + \frac{\mu_i}{L_i^2}\})$, the parameters of a neural network trained with quadratic line-search satisfies:
		\begin{equation}
			\mathbb{E}[\norm{\theta_{k}-\theta^*}^2]\leq M^k \norm{\theta_0-\theta^*}^2
		\end{equation}
		where $M=\mathbb{E}_{i}[\max\{g_{i}(\alpha_{max})\text{, }1-\frac{\mu_{i}}{L_{i}}\}]\in (0,1)$ is the convergence rate, and $g_{i}(\alpha_k) = 1-(\mu_{i}+L_{i})\alpha_k+L_{i}^2\alpha_k^2$.
		
		\label{theorem:convergence}
	\end{theorem}
	
	\subsubsection{Least Squares Estimation}\label{sec:ls}
	Notice that \cref{theorem:convergence} assumes the loss landscape to be strongly convex and $L$-smooth. In reality, we find that such conditions are satisfied easier in the later than earlier phase of training. In other words, the loss landscape is more rough at the beginning, and the 2-point method sometimes counters difficulties getting a good estimation $q(\alpha)$. 
	For instance, in rough surfaces like \cref{fig:rough_est}, the 2-point method may focus too much on the local information at the origin (both zero's order and first order moments.), which may lead to poor estimation.
	This motivates us to further develop a more robust version of the quadratic line-search using least squares estimation. 
	
	Instead of only using two samples (plus gradient), the Least Squares estimator takes more samples. And instead of finding a perfect match on the samples, Least Squares minimizes the mean squared error among all samples. 
	In our case, Least Squares minimizes the following objective:
	\begin{equation}
		\min_\omega \norm{X\cdot\omega-Y}_2^2=\frac{1}{n}\sum_i (A\alpha_i^2+B\alpha_i+C-\phi(\alpha_i))^2
	\end{equation}
	where 
	\begin{equation}
		X=
		\begin{bmatrix}
			\alpha_1^2 & \alpha_1 & 1\\
			\alpha_2^2 & \alpha_2 & 1\\
			& ...
		\end{bmatrix}
		\in \mathds{R}^{n\times 3}
	\end{equation} 
	is the sample matrix formulated by each step $\alpha_i$, $\omega=[A, B, C]^T$ are the coefficients to be optimized, and $Y=[\phi(\alpha_1), \phi(\alpha_2),...]^T\in\mathds{R}^n$ is the observation vector. The Least Squares estimator is: $\omega=(X^T X)^{-1} X^T Y$, where $(X^T X)^{-1}X^T$ is known as the Moore-Penrose inverse.
	
	\begin{algorithm}
		\small
		\caption{Quadratic line-search (Least Squares)}\label{alg:qlsls}
		\begin{algorithmic}[1]
			\Procedure {\texttt{QLS2}}{$f(\cdot), \Vec{\theta}_t, \Vec{p_t}, K_{max},\alpha_{max},slope$}
			\State {Initialize: $\gamma$(sample distance); $\alpha_0$; }
			\State {$X=\{(0,1,0),(0,0,1)\}$; $Y=\{slope,f(\theta_t) \}$}
			\For{$k \gets 1$ to $K_{max}$}
			\State{$x_k\gets (\gamma^2, \gamma, 1)$}
			\State{$X\gets X\bigcup\{x_k\}$}
			\State{$Y\gets Y\bigcup\{f(\theta_t+\gamma\cdot~p_t)\}$}
			\State{$q_{A,B,C}(\cdot)\gets \texttt{LeastSquare}(X,Y)$}
			\If{$A<=0$}
			\State{\textbf{return} $\gamma$}
			\ElsIf{$q_{ABC}(\alpha^*)>0 ~\&~ f(\theta_t+\alpha^* p_t)<f(\theta_t)$}
			\State{\textbf{return} $\alpha^*$}
			\Else{}
			\State{$\gamma\gets\alpha^*$}
			\EndIf
			\EndFor
			\State \textbf{return} $\alpha^*$
			\EndProcedure
		\end{algorithmic}
	\end{algorithm}

	The algorithm is described in \Cref{alg:qlsls}. When the loss landscape $\phi(\alpha)$ is rough, as shown in \Cref{fig:rough_est}, the slope of one point is not useful in order to capture the landscape trend. By using least squares estimation on multiple sample points without using slope, the quadratic interpolation becomes insensitive to gradient information near the origin, and seeks to accommodate more global information. 
	Through experiments in \Cref{sec:exp}, we show that using least squares is more robust.
	
	\subsection{Dynamically Adjusting Momentum}\label{sec:polak}
	In this section, we propose an automatic computation of the momentum factor. In SGD or its variants, this is manually set and fixed in value. We build a bridge between the momentum in SGD and the conjugate gradient parameter $\beta_t$. 
	(Recall that in the Nonlinear Conjugate Gradient (CG) method, the conjugate direction $s_t$ at iteration $t$ is updated as $s_t=-\nabla f+\beta_t\cdot s_{t-1}$, and parameters are updated by $x_{t+1}=x_t+\alpha_t\cdot s_t$.)
	$\beta_t$ has a deterministic optimal value (s.t. $s_t$ is conjugate to $s_{t-1}$) using the Hessian matrix in a quadratic objective. But for a general nonlinear case, various heuristic formulas are proposed~\cite{hestenes1952,fletcher1964,polak1969,dai1999,shewchuk1994}. $\alpha_t$ can be determined using either the Newton–Raphson method or a line-search such as the Armijo rule~\cite{armijo1966} and Wolfe conditions~\cite{wolfe69,wolfe71}. 
	The Nonlinear CG method generalizes various gradient-based optimization methods. 
	For instance, when $\beta_t=0$ and $\alpha_t$ is constant, CG degenerates as the Gradient Descent,and when $\beta_t$ is a non-zero constant, it instantiates the Gradient Descent with momentum, which is also discussed in \cite{Bhaya2004}. Moreover, the CG framework can also be used in the stochastic settings, where gradients $\nabla_\theta c(\theta)$ are replaced by partial gradients $\nabla_\theta~c_{batch}(\theta)$ in each iteration.
	
	With our proposed Quadratic line-search, the local solver is able to adaptively change the learning rate for performance. We next show that using the CG framework, the momentum term in SGD ($\beta_t$ in CG), which is usually constant in a SGD solver, can also be adaptively adjusted.
	
	For a quadratic objective of the following form:
	\begin{equation}
		\min f(\theta)=\frac{1}{2}\theta^TQ\theta+b^T\theta
	\end{equation}
	where $\theta\in\mathds{R}^n$ is the model parameters, and $Q\in\mathds{R}^{n\times n}$ and $b\in\mathds{R}^n$ are problem-dependent matrices. The optimal $\beta_t$ can be chosen as
	\begin{equation}
		\beta_t=\frac{\nabla f(\theta_t)^T\cdot Q \cdot p_{t-1}}{p_{t-1}^T Q p_{t-1}}
	\end{equation}
	where $p_{t-1}$ is the update direction at time $t-1$. Then the next update direction $p_t=-\nabla f(\theta_t)+\beta_t\cdot p_{t-1}$ is guaranteed to be $Q$-conjugate to $p_{t-1}$, which accelerates training by eliminating influences on previous progress.
	For general nonlinear objective functions with no explicit $Q$ and the Hessian is difficult to compute, different heuristics were proposed in \cite{hestenes1952,fletcher1964,polak1969,dai1999,shewchuk1994}. Since the classic Gradient Descent with momentum method is a stationary version of the Conjugate Gradient~\cite{Bhaya2004}, a dynamic choice of $\beta_t$, instead of fixed, can potentially speed up the training process. It is also notable that to compute $\beta_t$ using the aforementioned formula, no Hessian is required. For example, the Polak-Ribiere~\cite{polak1969} formula:
	\begin{equation}
		\beta^{PR}_t=\frac{\nabla f(\theta_t)^T(\nabla f(\theta_t)-\nabla f(\theta_{t-1}))}{\nabla f(\theta_{t-1})^T\nabla f(\theta_{t-1})}
	\end{equation}
	only requires the recent two gradients.
	
	A bounded version of the Polak-Ribiere momentum is used, \ie, to compute the momentum as $\beta'_t=\min\{\max\{0,\beta_t^{PR}\},\beta_{max}\}$.
	Through experiments, it is shown that by applying such dynamic momentum $\beta'_t$, the performance is comparable to or even better than the best manual momentum. The major advantage is that another crucial hyperparameter is eliminated from hand tuning.
	
	\subsection{Optimizations for Large Datasets}\label{sec:effi}
	\begin{wrapfigure}{r}{0.5\textwidth}
		\centering
		\includegraphics[width=\linewidth]{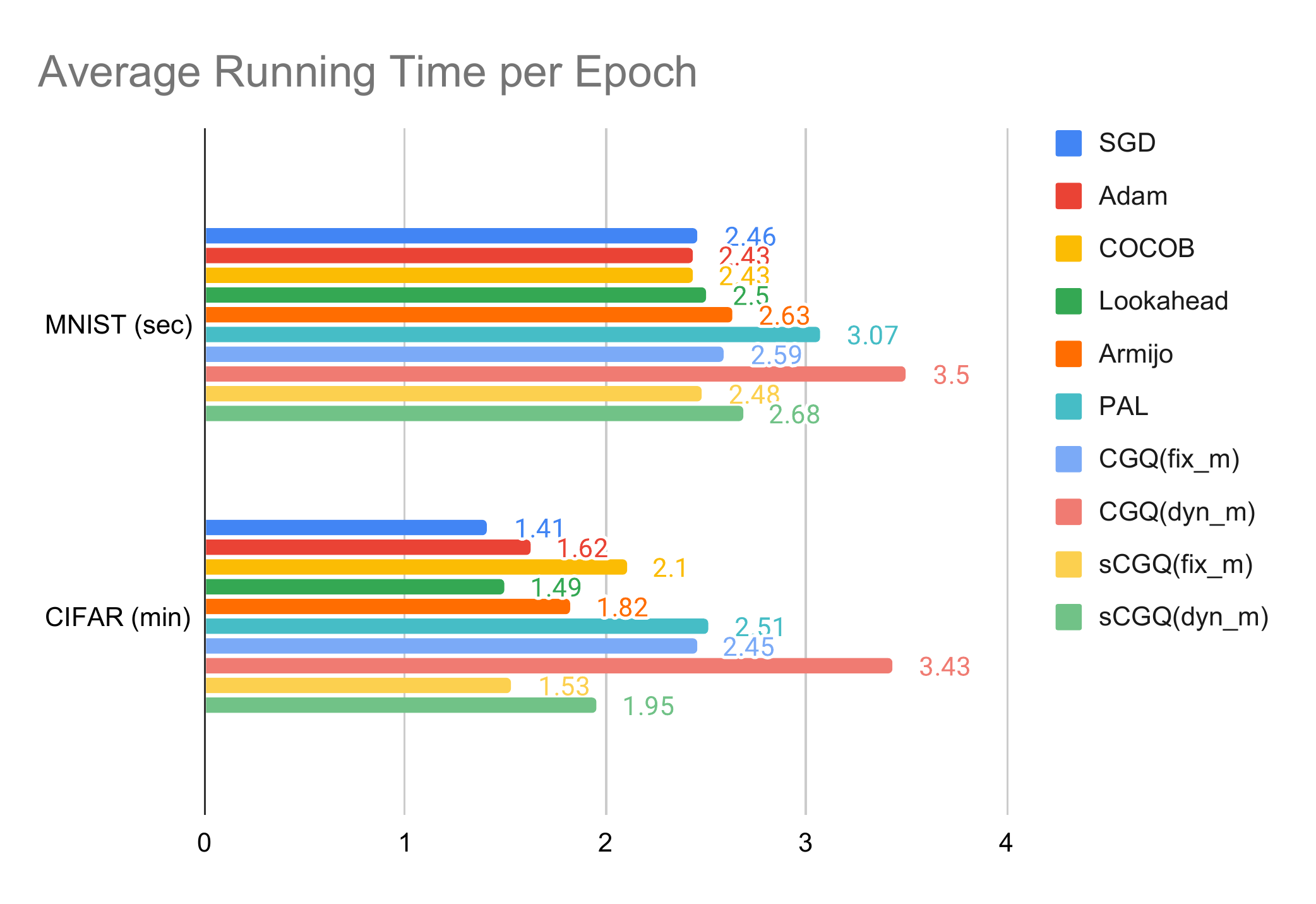}
		\caption{\small Running time comparison. sCGQ: stochastic line search with probability 0.1. \texttt{fix\_m}: fixed momentum; \texttt{dyn\_m}: Polak-Ribiere momentum. 2-point interpolation and Least Squares takes similar time.}
		\label{fig:epoch_time}
	\end{wrapfigure}
	Conceptually, CGQ performs a line search at every iteration. This unsurprisingly introduces more computation. To alleviate this issue on large systems, 
	it is reasonable to use partially observed curvature (from batches) to estimate that of the whole dataset. This motivates us to implement the sCGQ that performs line search stochastically: For each training batch, the optimizer will perform a line search (that optimizes learning rate) with a certain probability $p$. For batches without line searches, the moving average of past line search results will be applied. \cref{fig:epoch_time} show that this makes the run time of sCGQ with fixed momentum comparative to naive SGD, and sCGQ with dynamic momentum comparative to other line serach methods. Experiments in \cref{sec:ablation_prob} and the runtime comparison in \cref{fig:epoch_time} show that sCGQ benefits from both fast execution and good estimation of 1-D loss landscape (in terms of learning rate).
	
	\section{Experiments}\label{sec:exp}
	To empirically evaluate the performance of the proposed method, we design a thorough experiment consisting of two parts described in the following sections. We focus on comparing our method with different configurations in \Cref{sec:ablation}. In \Cref{sec:imagedata}, we benchmark our method with six popular optimization methods: SGD+momentum, Adam~\cite{adam}, COCOB~\cite{cocob}, Lookahead optimizer~\cite{lookahead}, SGD+Armijo Line-Search~\cite{painless}, and PAL~\cite{pal2020}. 
	We manually tune their hyperparameters and the results summarized in the following sections are their best performances in the experiments to our best knowledge.
	Specifically for SGD and ADAM, the learning rates are fixed (SGD: 0.02 for VGG, 0.01 on MNIST, and 0.1 elsewhere; Adam: 0.001). 
	There is no learning rate for COCOB.
	For our own method, we tested four varients: CGQ(2pt)/CGQ(LS): complete line search with quadratic interpolation or least squares estimation; sCGQ(2pt)/sCGQ(LS): stochastic line search with quadratic interpolation/least squares estimation.
	We showcase the fast convergence of our method on training multi-class classifiers for widely used benchmark image datasets including MNIST, SVHN, CIFAR-10, and CIFAR-100. For fairness, model parameters are initialized the same among all solvers for each test case, and we run multiple times and show the average performance for each case. All experiments are conducted on a NVIDIA GTX-1080Ti GPU, and the program is developped under PyTorch 1.7.0 framework. 
	Code is available in \url{https://github.com/zh272/CGQ-solver}.
	
	\begin{table*}[t]
		\centering
		\small
		\begin{tabular}{ l | c | c | c | c }
			\toprule \hline
			\multirow{3}{*}{\textbf{Method}} & \multicolumn{4}{c}{\textbf{Train Loss ~/~ Test Accuracy}}\\
			\cline{2-5}
			& MNIST                         & \multicolumn{3}{c}{SVHN} \\
			\cline{2-5}
			& MLP                       & VGG-16                & ResNet-110                & DenseNet-100 \\
			\hline
			SGD                         & .076 / 97.24             & .080 / 93.16           & .217 / 92.68             & .158 / 94.34 \\
			Adam~\cite{adam}            & .014 / 98.02             & .026 / 93.14           & .134 / 95.34             & .093 / 95.35 \\
			Lookahead~\cite{lookahead}  & .131 / 95.99             & .039 / 93.95           & .157 / 93.14             & .112 / 94.12 \\
			COCOB~\cite{cocob}          & .267 / 92.72             & .036 / 91.81           & .069 / 90.34             & .035 / 91.30 \\
			Armijo~\cite{painless}      & .023 / 97.85             & .008 / 92.42           & --                       & -- \\
			PAL~\cite{pal2020}          & .021 / 97.60             & --                     & .016 / 94.73             & .017 / 95.10 \\
			\hline
			CGQ(2pt)                    & \highlight{.007/98.24}  & .011 / 94.70             &\highlight{.009}/ 95.31   & \highlight{.004/96.11} \\
			CGQ(LS)                     & \highlight{.007/98.24}  & \highlight{.006}/ 94.79  & .012 / 95.81             &  .008 / 95.99 \\
			sCGQ(2pt)                   & .008 /\highlight{98.24}  & .020 / 93.89            & .043 / \highlight{95.84} & .029 / 94.23 \\
			sCGQ(LS)                    & .008 / 98.12             & .016 / \highlight{94.93}& .039 / 95.63 & .032 / 95.43 \\
			\hline \bottomrule
		\end{tabular}
		\caption{\small Performance comparison of different optimization algorithms on the MNIST and SVHN datasets. Training budget: 20 epochs (MNIST) / 40 epochs (SVHN). ``-" means not converging.}
		\label{tab:mnist_svhn}
	\end{table*}
	
	\subsection{Ablation Test}\label{sec:ablation}
	
	\subsubsection{Dynamical Momentum}\label{sec:ablation_m}
	Applying SGD with momentum to reach faster convergence has become common in deep neural network training. Heavy ball~\cite{polyak} and Nesterov momentum~\cite{nesterov} are two classic static acceleration scheme that consider the momentum term as a fixed hyperparameter. In this section, we compare the performances of SGD with quadratic line-search plus (1) Heavy ball momentum\cite{polyak} with coefficients from 0.1 to 0.9, (2) Fletcher-Reeves momentum \cite{fletcher1964}, (3) Polak-Ribiere momentum \cite{polak1969}, (4) Hestenes-Stiefel momentum \cite{hestenes1952}, and (5) Dai-Yuan momentum \cite{dai1999}.
	
	\begin{figure*}[t]
		\centering
		\begin{subfigure}[b]{0.48\textwidth}
			\includegraphics[width=\textwidth]{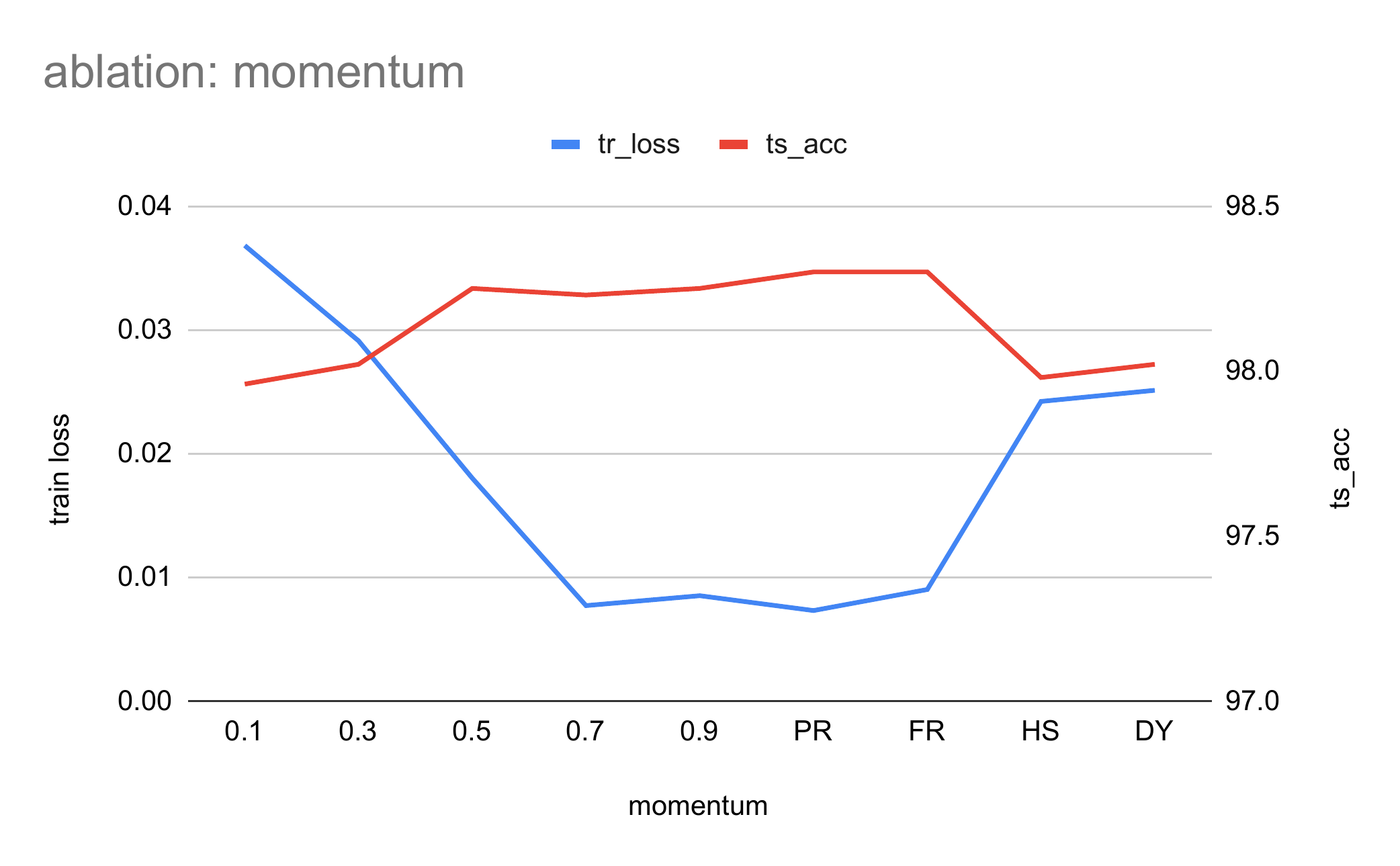}
			\caption{}
			\label{fig:ablation_mom}
		\end{subfigure}%
		\hfil
		\begin{subfigure}[b]{0.49\textwidth}
			\includegraphics[width=\textwidth]{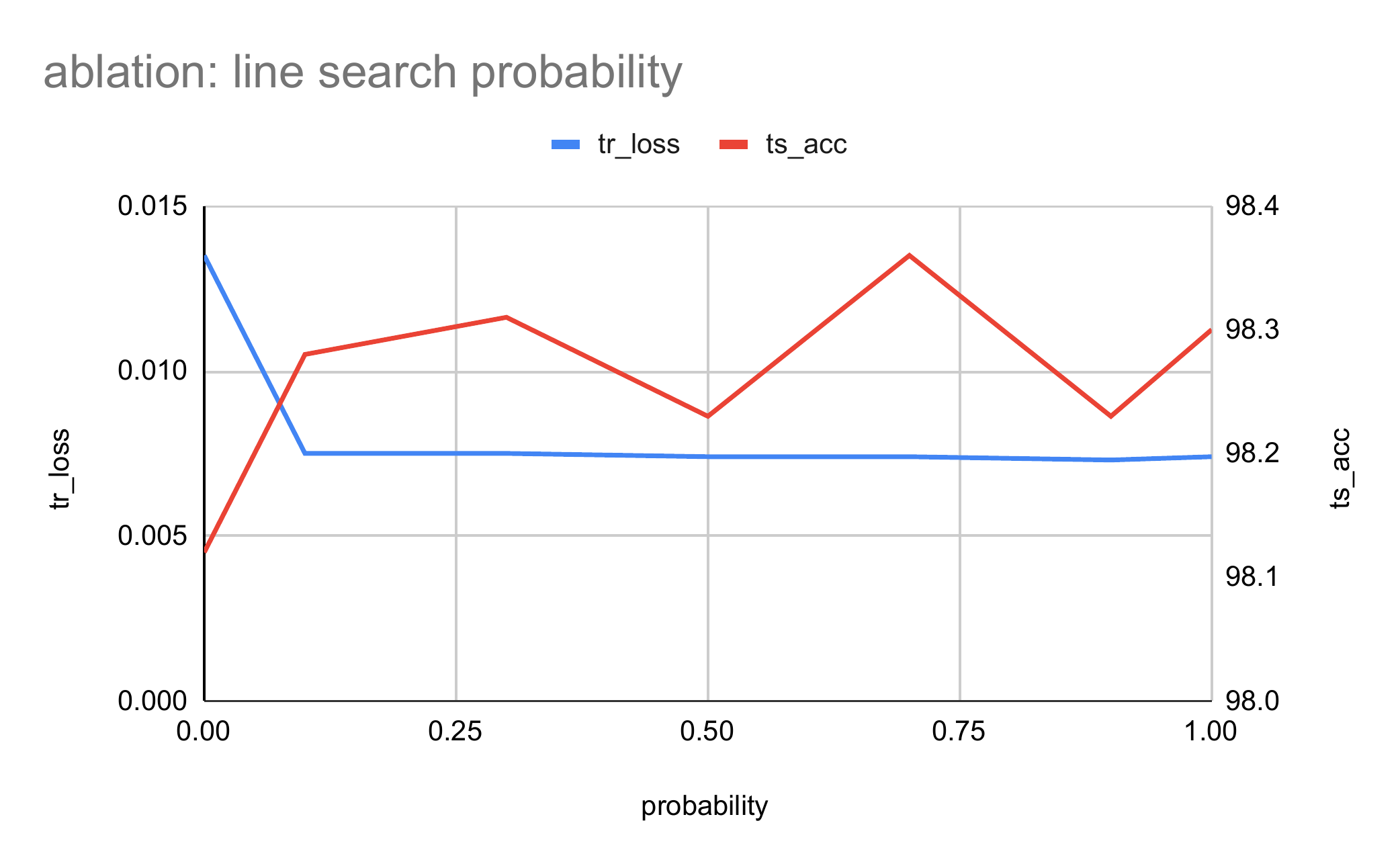}
			\caption{}
			\label{fig:ablation_lsprob}
		\end{subfigure}
		\hfil
		\begin{subfigure}[b]{0.6\textwidth}
			\includegraphics[width=\textwidth]{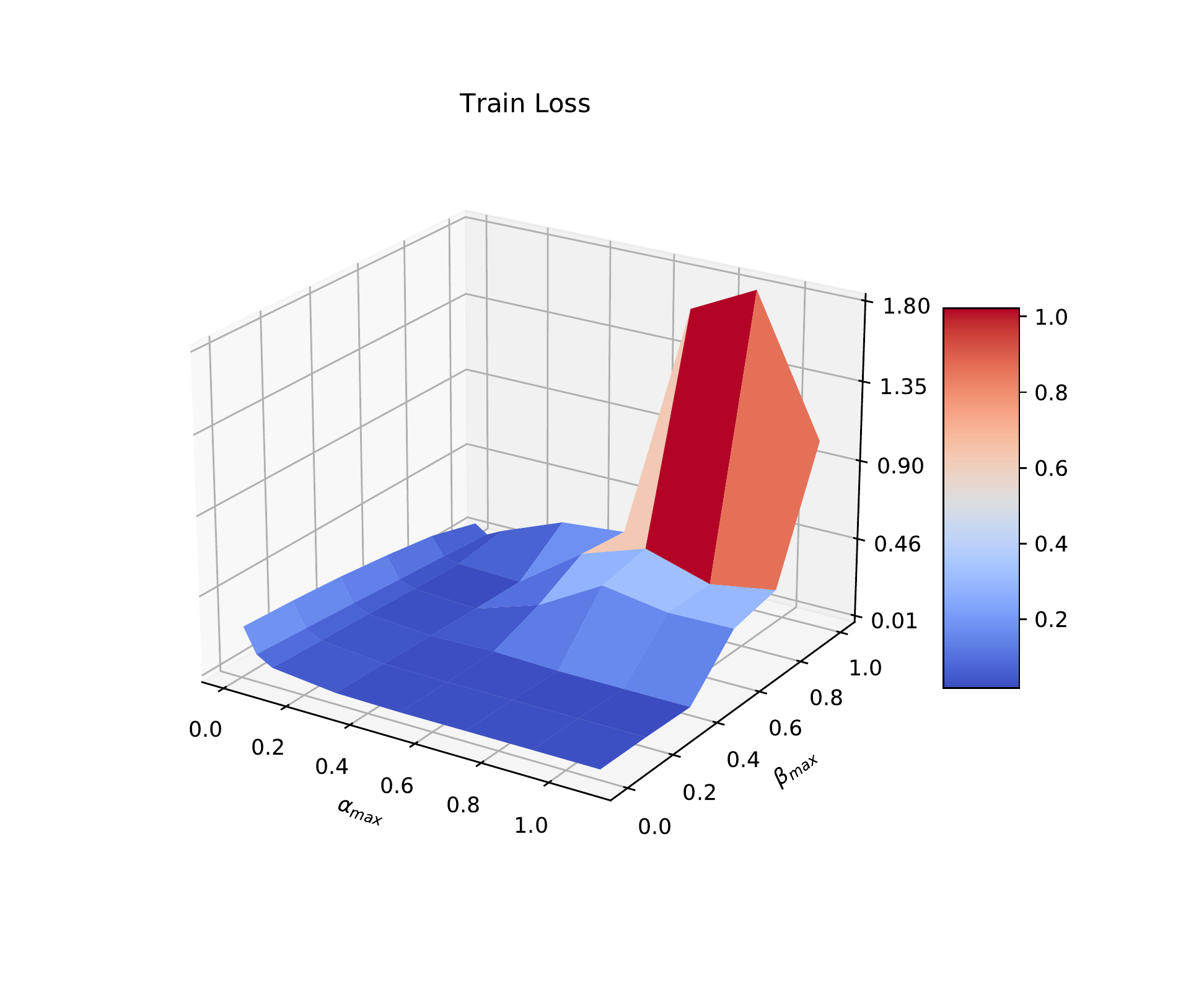}
			\caption{}
			\label{fig:ablation_alpha_beta}
		\end{subfigure}
		\caption{\small Training loss (One-layer MLP trained on MNIST dataset) w.r.t.: (a) Different momentum settings. (b) Line search probability. ``0": naive SGD with no line search; ``1": line search on every mini-batch. (c) $\alpha_{max}$ and $\beta_{max}$.}
		\label{fig:ablation}
	\end{figure*}
	The benchmark experiment is run on the MNIST dataset. The comparison results are summarized in \Cref{fig:ablation_mom}. On the basis of the results of the ablation test, it is promising to use Polak-Ribiere momentum as a dynamic acceleration scheme for our method so that the momentum term is no longer a hyperparameter that needs tuning manually. 
	
	\subsubsection{Line Search Probability}\label{sec:ablation_prob}
	To evaluate the impact of the randomness on line search, we test our method with probability thresholds $p$ varying from 0(no line search) to 1(complete line search). Results in \cref{fig:ablation_lsprob} show that line search improves the final model performance even with small probability $p$, and it does not yield further improvements as $p$ increases. 
	
	\subsubsection{$\alpha_{max}$ and $\beta_{max}$}\label{sec:ablation_bound}
	$\alpha_{max}$ (upper bound of learning rate) and $\beta_{max} $ (upper bound of momentum), are two important hyperparameters in CGQ. In this section we perform an ablation test to show the feasible regions. We perform a grid search on the 2D space by $\alpha_{max}$ (0.01 to 1.1) and $\beta_{max}$ (0 to 1). It is observed from \Cref{fig:ablation_alpha_beta} that the model performs well when $\alpha_{max}$ and $\beta_{max}$ are not both large. And the model will diverge when $\beta_{max}$ is larger than 1 (thus not shown on figure). Empirically good choices are: $\beta_{max}=0.8$; $\alpha_{max}=0.3$ (ResNet) / 0.1 (DenseNet) / 0.05 (VGG).

	\begin{table*}[t]
		\centering
		\small
		\begin{tabular}{ l | c | c | c | c | c | c }
			\toprule \hline
			\multirow{3}{*}{\textbf{Method}} & \multicolumn{6}{c}{\textbf{Train Loss ~/~ Test Accuracy}}\\
			\cline{2-7}
			& \multicolumn{2}{c|}{VGG-16}                           & \multicolumn{2}{c|}{ResNet-164}                   & \multicolumn{2}{c}{DenseNet-100} \\
			\cline{2-7}
			& CIFAR-10                          & CIFAR-100             & CIFAR-10                  & CIFAR-100                 & CIFAR-10              & CIFAR-100 \\
			\hline
			SGD & .116 / \highlight{89.64}              & .685 / 64.37          & .271 / 86.06              & .860 / 63.55             & .255 / 86.71           & .901 / 62.06  \\
			Adam & .044 / 88.06                         & .408 / 55.21          & .144 / 90.06              & .419 / 68.07             & .127 / 90.42           & .364 / 68.23 \\
			\tiny{Lookahead} & .069 / 89.21             & .369 / 60.97          & .155 / 88.33              & .526 / 59.21             & .153 / 88.50           & .594 / 61.62 \\
			\scriptsize{COCOB} & .087 / 84.78           & .772 / 44.83          & .330 / 78.96              & 1.450/ 48.00             & .205 / 80.63           & 1.103/ 51.56 \\
			Armijo & .298 / 81.18                       & .025 / 62.50          & \highlight{.003}/ 93.64   & \highlight{.004}/ 73.26  & .009 / 92.72           & .086 / 65.55 \\
			PAL & --                                    & --                    & .017 / 92.04              & .452 /  64.01            & .023 / 90.99           & .072 / 65.90 \\
			\hline
			\tiny{CGQ(2pt)} & \highlight{.008}/ 88.21  & .025 / 65.46          &  .007 / 93.04             & .014 / 75.26             & \highlight{.009}/ 91.49 & \highlight{.020}/ 75.01 \\
			\tiny{CGQ(LS)} & .010 / 88.64              & \highlight{.019/66.49}& .010 / 93.52              & .019 / 73.43             & .010 / 92.23            & .029 / 74.35 \\
			\tiny{sCGQ(2pt)} & .035 /89.16             & .038 / 63.12          & .017 /\highlight{93.75}   & .016 /\highlight{75.94}   & .020 / \highlight{93.71}& .035 /\highlight{76.15} \\
			\tiny{sCGQ(LS)} & .041 /  88.88            & .044 / 62.93          & .016 / 93.56              & .016 / 75.60              & .028 /93.13             & .032 / 76.04 \\
			\hline \bottomrule
		\end{tabular}
		\caption{\small Performance comparison for the CIFAR-10 and CIFAR-100 datasets. Training budget: 200 epochs. ``-" means the method does not converge.}
		\label{tab:cifar}
	\end{table*}
	
	\subsection{Multi-class classification on image datasets}\label{sec:imagedata}
	Recent evidence has shown that deep convolutional neural networks lead to impressive breakthroughs in multi-class image classification. We mainly select the three popular families of architectures of deep convolutional neural networks: ResNet, DenseNet and VGG to benchmark our method with the other powerful optimizers. The comparison results will be presented in the following two sections, according to different datasets used in experiments.
	
	\subsubsection{MNIST and SVHN}
	The MNIST dataset contains 60,000 hand-written digit images for training, and 10,000 images for testing. Images in MNIST are grayscale images with size $28 \times 28$. SVHN dataset consists of 73,257 images in the training set, 26032 images in the testing set and another 531,131 additional images provided for training. All the images in SVHN are colored digit images with a size of $32\times32$. For MNIST, we construct a fully connected network with one hidden layer of 1000 neurons as a toy benchmark example. For SVHN, we experiment with three more expressive architectures: VGG-16, ResNet-110, and DenseNet-100.
	
	The main results are summarized in \Cref{tab:mnist_svhn}. 
	We observed that with the same training budget, CGQ presents better performances in terms of both training loss and test accuracy. The stochastic version of 
	We omit SGD with Armijo-line-search in the SVHN experiment since it appeared to be unstable and easy to diverge.
	
	\subsubsection{CIFAR-10 and CIFAR-100}
	\begin{figure}[t]
		\centering
		\begin{subfigure}[b]{0.49\textwidth}
			\includegraphics[width=\linewidth,height=0.8\linewidth]{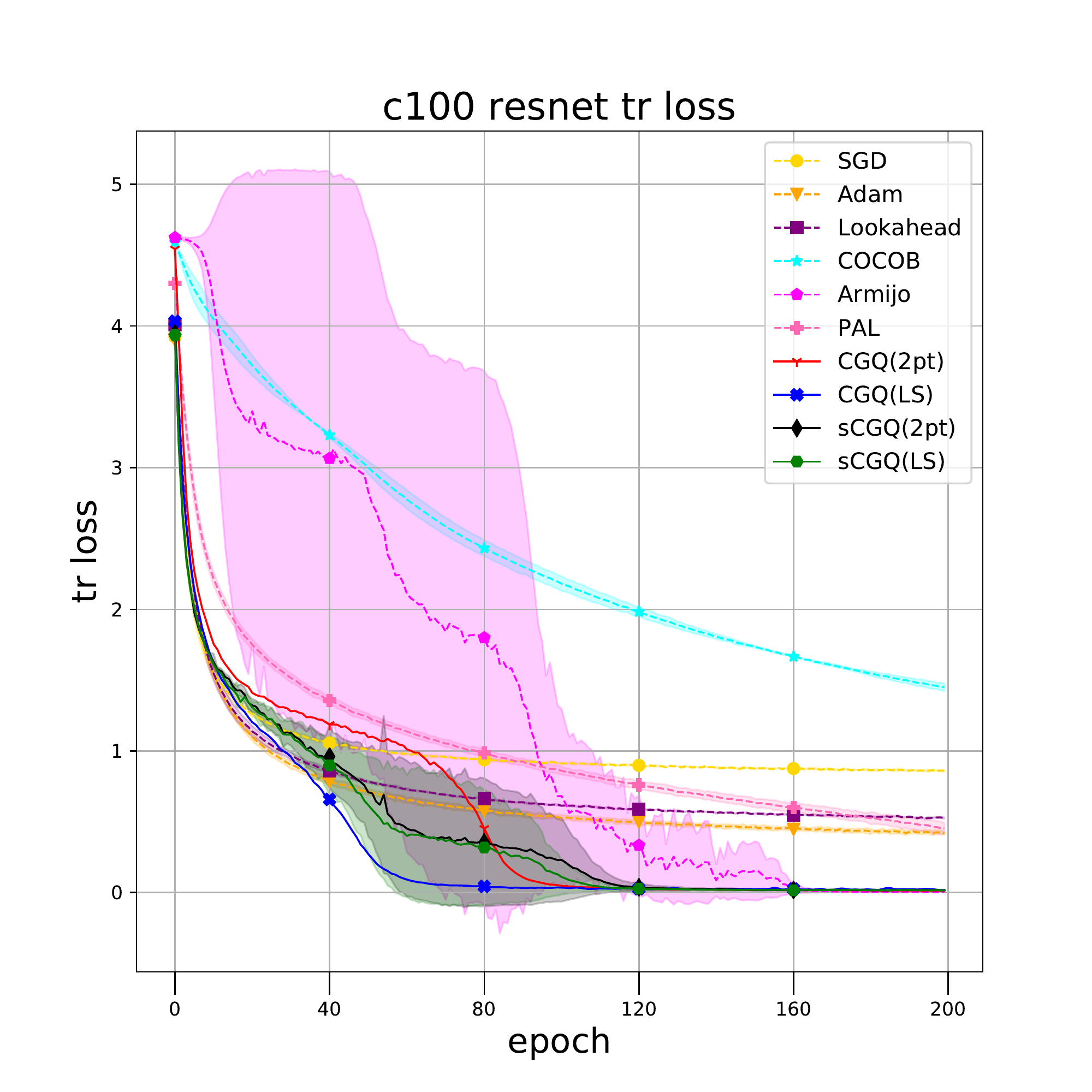}
			\caption{}
		\end{subfigure}
		\hfil
		\begin{subfigure}[b]{0.49\textwidth}
			\includegraphics[width=\linewidth,height=0.8\linewidth]{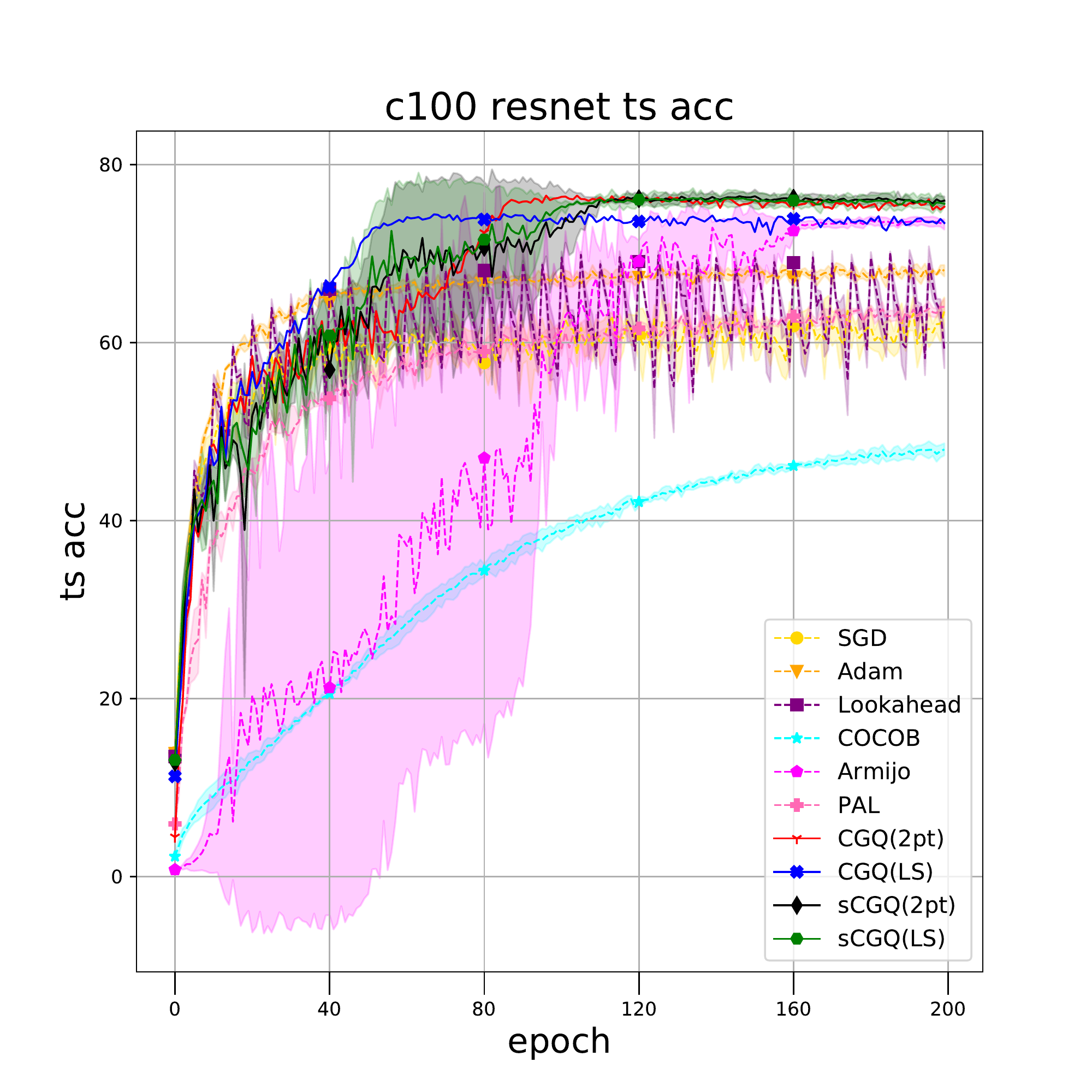}
			\caption{}
		\end{subfigure}
		\hfil
		\begin{subfigure}[b]{0.49\textwidth}
			\includegraphics[width=\linewidth,height=0.8\linewidth]{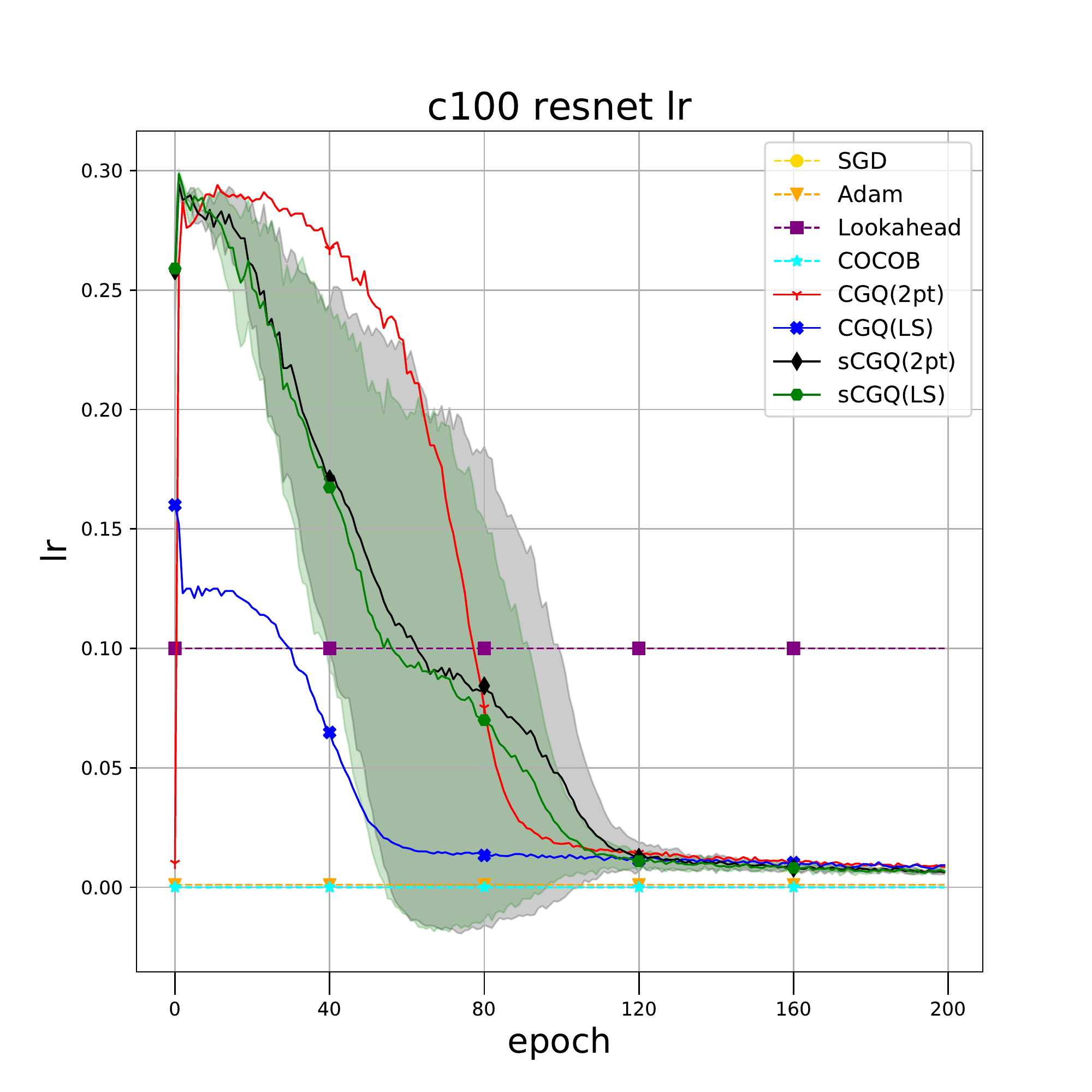}
			\caption{}
		\end{subfigure}
		\hfil
		\begin{subfigure}[b]{0.49\textwidth}
			\includegraphics[width=\linewidth,height=0.8\linewidth]{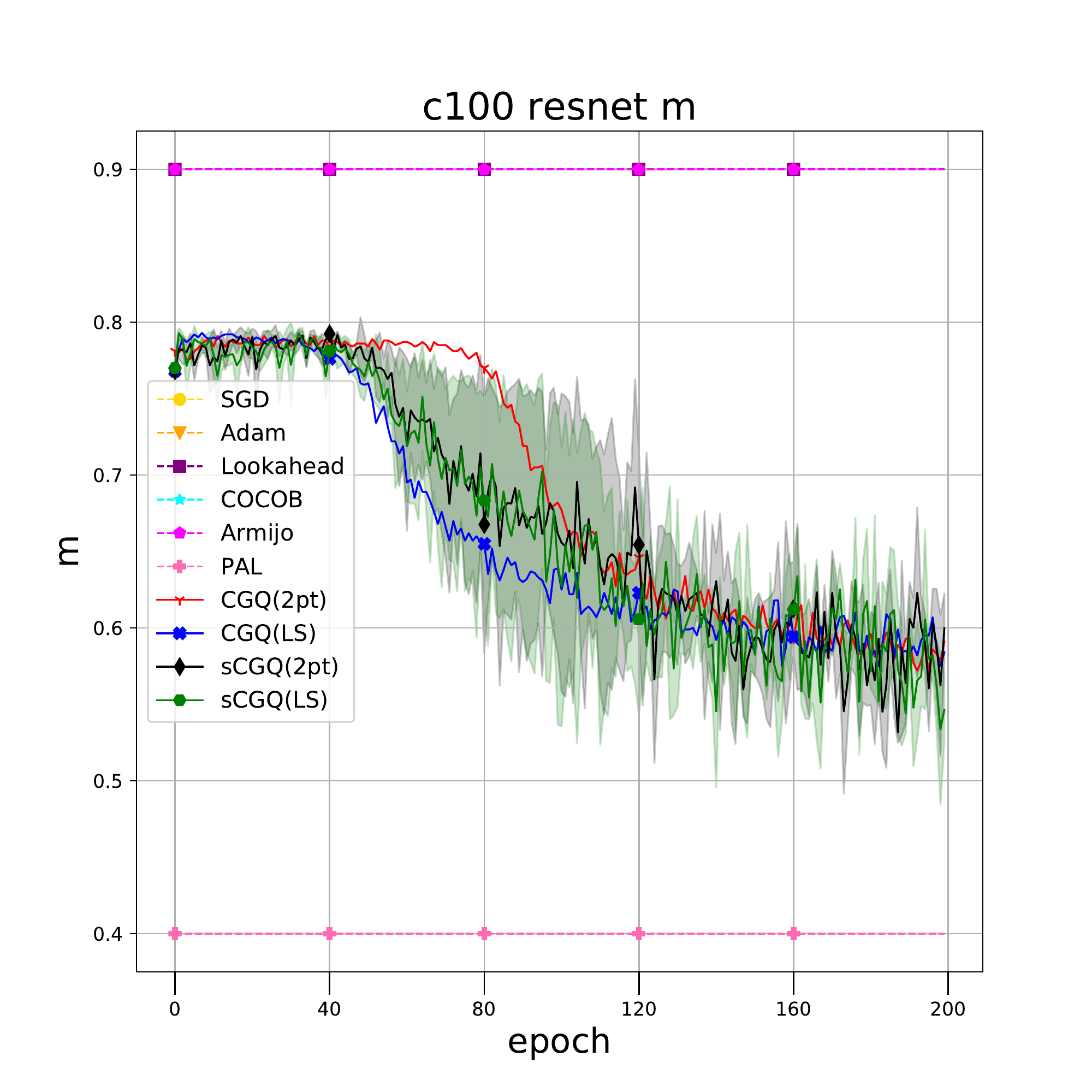}
			\caption{}
		\end{subfigure}
		\caption{\small Dynamics of training a ResNet-164 with various optimization methods on the CIFAR100. (a) Training Loss. (b) Test Accuracy. (c) Learning Rate. (d) Momentum}
		\label{fig:cgq_better}
	\end{figure}
	The CIFAR-10 and CIFAR-100 datasets both consist of 50,000 images for training and 10,000 images for testing, while images are in 10 and 100 different classes, respectively. We run the experiments using VGG-16, ResNet-164, and DenseNet-100. Each of model-algorithm pair is trained for 200 epochs with same initialization points. 
	
	In \Cref{tab:cifar}, we record the training loss and test accuracy of each model and optimizer pair in the CIFAR-10 and CIFAR-100 experiments. 
	In most of the cases, CGQ and sCGQ not only reach the best performance among all line search methods, but also converges to a local minimum faster. In the the more challenging CIFAR-100 case, the margin is more significant. \cref{fig:cgq_better} shows the dynamics for training ResNet on the CIFAR100 dataset. Each line plots the mean over multiple runs, while corresponding standard deviations are shown as shaded area around the mean. 
	(a) and (b) tells that CGQ variants performs stably better in training loss and testing accuracy compared to other compared methods.
	Moreover, (c) and (d) demonstrates the automatic decaying effect on learning rate and momentum of the CGQ method.
	
	It is observed from the tables that CGQ performs better than sCGQ on the training set, but sCGQ has better test time performance. In this sense,
	the sCGQ is preferable to CGQ for larger datasets, not only because of faster execution time, but also because of better generalization. 
	It is also observed that 2-point interpolation method performs better than its Lease Squares variant on larger datasets (CIFAR). This could due to an underestimation of the smoothness for deep neural networks. This is also validated through a visualization of the loss landscape along the line search direction in \cref{fig:landscape}.
	
	\section{Conclusion}
	In this paper, we propose CGQ method to replace hand tuning two crucial hyperparameters in SGD, \ie, learning rate and momentum. The learning rate is determined by our quadratic line-search method, and momentum is adaptively computed using the bounded Polak-Ribiere formula. 
	
	Experiments on modern convolutional neural networks show that CGQ converges faster due to its adaptiveness to the loss landscape. Theoretical results are also provided towards the convergence of quadratic line search under convex and smooth assumptions.
	To improve efficiency on larger models, we further propose the sCGQ method that performs line search on fewer iterations. This modification improves not only the run time, but also the generalization capability in terms of test accuracy. 
	In most cases, the CGQ method outperforms other local methods.
	
	\bibliographystyle{splncs04}
	\bibliography{references}

	\appendix
	\section{Proof of Lemma 1}\label{app:proof_lemma1}
	Suppose $h_{ik}(\alpha)=f_{ik}(\theta_k-\alpha \nabla f_{ik}(\theta_k))$. Besides $\alpha=0$, sample the other point $\alpha=\alpha_{1}$ and get the quadratic interpolant of $f_{ik}$ as follows,
	
	\[q(\alpha)=\frac{\norm{\nabla f_{ik}(\theta_k)}^2\alpha_1 + h_{ik}(\alpha_1)-h_{ik}(0)}{\alpha_1^2}\alpha^2-\norm{\nabla f_{ik}(\theta_k)}^2\alpha + h_{ik}(0).\]
	Thus, the step size $\alpha_k$ returned by a quadratic line-search is
	\[\alpha_k=\frac{\norm{\nabla f_{ik}(\theta_k)}^2\alpha_1^2}{2(\norm{\nabla f_{ik}(\theta_k)}^2\alpha_1 + h_{ik}(\alpha_1)-h_{ik}(0))}.\]
	According to the assumption on the smoothness of $f_{ik}$ and the stochastic gradient descent update rule, the following inequality holds: 
	\[h_{ik}(\alpha_1)\leq h_{ik}(0) + \nabla f_{ik}(\theta_k)'(-\nabla f_{ik}(\theta_k)\alpha_1) + \frac{L_{ik}}{2}\norm{-f_{ik}(\theta_k)\alpha_1}^2.\]
	Thus,
	\[h_{ik}(\alpha_1)-h_{ik}(0)\leq (-\alpha_1 + \frac{L_{ik}}{2}\alpha_1^2)\norm{\nabla f_{ik}(\theta_k)}^2.\]
	As a result,
	\[\alpha_k \geq \frac{\norm{\nabla f_{ik}(\theta_k)}^2\alpha_1^2}{2(\norm{\nabla f_{ik}(\theta_k)}^2\alpha_1 + (-\alpha_1 + \frac{L_{ik}}{2}\alpha_1^2)\norm{\nabla f_{ik}(\theta_k)}^2)}=\frac{1}{L_{ik}}.\]
	Similarly, based on the assumption on the strong convexity of $f_{ik}$ and the stochastic gradient descent update rule,
	\[h_{ik}(\alpha_1)-h_{ik}(0)\geq (-\alpha_1 + \frac{\mu_{ik}}{2}\alpha_1^2)\norm{\nabla f_{ik}(\theta_k)}^2.\]
	Therefore,
	\[\alpha_k \leq\frac{1}{\mu_{ik}}.\]
	By definition, $\alpha_k \in (0, \alpha_{max}]$, and thus, the step size returned by quadratic line-search satisfies the following relation:
	\[\alpha_k \in [min\{\frac{1}{L_{ik}}\text{, }\alpha_{max}\}\text{, } min\{\frac{1}{\mu_{ik}}\text{, }\alpha_{max}\}].\] 
	\qed
	
	\section{Proof of Theorem 1}\label{app:proof_theorem1}
	Suppose the loss function $f(\theta)$ is minimized at $\theta^*$. According to the stochastic gradient descent update rule,
	\[\norm{\theta_{k+1}-\theta^*}^2 = \norm{\theta_k-\alpha_k\nabla f_{ik}(\theta_k)-\theta^*}^2=\norm{\theta_k-\theta^*}^2-2\alpha_k\langle \nabla f_{ik}(\theta_k) \text{, } \theta_k-\theta^*\rangle + \alpha_k^2\norm{\nabla f_{ik}(\theta_k)}^2.\]
	Based on the assumption on the strong convexity of $f_{ik}(\cdot)$,
	\[-\langle \nabla f_{ik}(\theta_k) \text{, } \theta_k-\theta^*\rangle \leq f_{ik}(\theta^*)-f_{ik}(\theta_k)-\frac{\mu_{ik}}{2}\norm{\theta_k - \theta^*}^2.\]
	Thus,
	\[\norm{\theta_{k+1}-\theta^*}^2 \leq (1-\mu_{ik}\alpha_k)\norm{\theta_k-\theta^*}^2+2\alpha_k[f_{ik}(\theta^*)-f_{ik}(\theta_k)]+ \alpha_k^2\norm{\nabla f_{ik}(\theta_k)}^2.\]
	Based on the assumption on the smoothness of $f_{ik}(\cdot)$,
	\[f_{ik}(\theta^*)-f_{ik}(\theta_k) \leq -\frac{1}{2L_{ik}}\norm{\nabla f_{ik}(\theta_k)}^2\]
	\[\norm{\nabla f_{ik}(\theta_k)-\nabla f_{ik}(\theta^*)}^2 \leq L_{ik}^2 \norm{\theta_k-\theta^*}^2.\]
	In the interpolation setting, $\nabla f_{ik}(\theta^*)=0$; therefore,
	\[\norm{\nabla f_{ik}(\theta_k)}^2 \leq L_{ik}^2 \norm{\theta_k-\theta^*}^2.\]
	As a result,
	\[\norm{\theta_{k+1}-\theta^*}^2 \leq [1-(\mu_{ik}+L_{ik})\alpha_k+L_{ik}^2\alpha_k^2]\norm{\theta_k-\theta^*}^2\]
	Let $g_{ik}(\alpha_k) = 1-(\mu_{ik}+L_{ik})\alpha_k+L_{ik}^2\alpha_k^2$. According to the assumptions, it is trivial to get $L_{ik} > \mu_{ik} > 0$, and thus,
	\[g_{ik}(\alpha_k) > 1-2L_{ik}\alpha_k+L_{ik}^2\alpha_k^2 \geq 0.\]
	For each $f_i$, define a constant $C_i = \frac{1}{L_i} + \frac{\mu_i}{L_i^2}$. If $\alpha_{max} \in (0, C_{min})$, $g_{ik}(\alpha_{max})<1$. Based on Lemma 1,
	\[0 \leq g_{ik}(\alpha_k) \leq max\{g_{ik}(\alpha_{max})\text{, } g_{ik}(\frac{1}{L_{ik}})\}= max\{g_{ik}(\alpha_{max})\text{, }1-\frac{\mu_{ik}}{L_{ik}}\}< 1.\]
	Therefore,
	\[\norm{\theta_{k+1}-\theta^*}^2 \leq max\{g_{ik}(\alpha_{max})\text{, }1-\frac{\mu_{ik}}{L_{ik}}\}\norm{\theta_k-\theta^*}^2.\]
	Take the expectation with respect to $i_k$:
	\[\mathbb{E}[\norm{\theta_{k+1}-\theta^*}^2]\leq \mathbb{E}_{ik}[max\{g_{ik}(\alpha_{max})\text{, }1-\frac{\mu_{ik}}{L_{ik}}\}]\norm{\theta_k-\theta^*}^2\]
	where $\mathbb{E}_{ik}[\max\{g_{ik}(\alpha_{max})\text{, }1-\frac{\mu_{ik}}{L_{ik}}\}] \in (0,~1)$.
	\qed
\end{document}